%% file: template.tex
\documentclass{article}

\usepackage{arxiv}

\usepackage[utf8]{inputenc} 
\usepackage[T1]{fontenc}    
\usepackage{hyperref}       
\usepackage{url}            
\usepackage{booktabs}       
\usepackage{amsfonts}       
\usepackage{nicefrac}       
\usepackage{microtype}      
\usepackage{lipsum}		
\usepackage{graphicx}
\usepackage{natbib}
\usepackage{doi}

\usepackage{amsmath}
\usepackage{multirow}
\usepackage{makecell}
\usepackage{bm}
\usepackage{wrapfig}
\usepackage{enumitem}
\usepackage{tabularray}
\usepackage{amssymb}
\usepackage{pifont}
\usepackage{fontawesome5}
\usepackage{marvosym}
\usepackage{siunitx}
\usepackage{algorithm, algorithmicx, algpseudocode}
\usepackage{threeparttable}

\usepackage{listings}
\usepackage{xcolor}

\lstdefinelanguage{json}{
  morestring=[b]",
  morecomment=[l]{//},
}

\title{RareAgent: Self-Evolving Reasoning for Drug Repurposing in Rare Diseases}

\author{
 \textbf{Lang Qin\textsuperscript{1}},
 \textbf{Zijian Gan\textsuperscript{2}},
 \textbf{Xu Cao\textsuperscript{3}},
 \textbf{Pengcheng Jiang\textsuperscript{3}},
\\
 \textbf{Yankai Jiang\textsuperscript{4}},
 \textbf{Jiawei Han\textsuperscript{3}},
 \textbf{Kaishun Wu\textsuperscript{1}},
 \textbf{Jintai Chen\textsuperscript{1}\thanks{Corresponding author. Email: \href{mailto:jintaichen@hkust-gz.edu.cn}{jintaichen@hkust-gz.edu.cn}}}\\
 \textsuperscript{1}	Information Hub, The Hong Kong University of Science and Technology (Guangzhou),\\
 \textsuperscript{2} College of Computer Science, Nankai University,\\
 \textsuperscript{3}	Department of Computer Science, University of Illinois at Urbana-Champaign,\\
 \textsuperscript{4} Shanghai Artificial Intelligence Laboratory
}

\newcommand{\model}{\textbf{RareAgent}}

\begin{document}
\maketitle

\setcounter{footnote}{-1}

\input{body/1_abstract}

\input{body/2_introduction}

\input{body/3_relatedwork}

\input{body/4_method}

\input{body/5_experiments}

\input{body/6_conclusion}

\bibliographystyle{unsrtnat}
\bibliography{references, custom}

\input{body/7_appendix}

\end{document}

%% file: body/1_abstract.tex
\begin{abstract}
Computational drug repurposing for rare diseases is especially challenging when no prior associations exist between drugs and target diseases. Therefore, knowledge graph completion and message-passing GNNs have little reliable signal to learn and propagate, resulting in poor performance.
We present RareAgent, a self-evolving multi-agent system that reframes this task from passive pattern recognition to active evidence-seeking reasoning.
RareAgent organizes task-specific adversarial debates in which agents dynamically construct evidence graphs from diverse perspectives to support, refute, or entail hypotheses.
The reasoning strategies are analyzed \textit{post hoc} in a self-evolutionary loop, producing textual feedback that refines agent policies,
while successful reasoning paths are distilled into transferable heuristics to accelerate future investigations.
Comprehensive evaluations reveal that RareAgent improves the indication AUPRC by 18.1\% over reasoning baselines and provides a transparent reasoning chain consistent with clinical evidence.

\end{abstract}

%% file: body/2_introduction.tex
\section{Introduction}

Drug repurposing, the discovery of new therapeutic uses for approved drugs, is pivotal for accelerating drug development while markedly reducing costs and clinical risks~\citep{liu_deep_2021, huang_foundation_2024, de_la_fuente_towards_2025}.
Its value is particularly pronounced in rare diseases, where fewer than 10\% of the thousands of identified rare diseases currently have approved therapies, leaving millions of patients with unmet medical needs~\citep{greene_genetic_2023, cummings_drug_2025, cipriani_rare_2025}.

Recent computational work on drug repurposing typically organizes biomedical evidence as knowledge graphs (KGs)~\citep{himmelstein_systematic_2017,drkg2020, zhang_drug_2021,  chandak_building_2023} and applies graph neural networks (GNNs) to infer new drug-disease associations~\citep{zitnik_modeling_2018, li_drug_2024}.
Yet their efficacy can be limited by the scarcity of curated \textbf{drug-disease} associations, especially when both the candidate drugs and the target diseases (\textit{e.g.}, rare diseases) lack reliable knowledge.
In practice, the \emph{drug side} is inherently \emph{sparse} as a drug typically has only a few officially recognized indications, while the \emph{disease side} becomes a \emph{degree-zero} node in the case where our target is a rare disease for which no drug-disease associations have been established within the current body of human knowledge.
We refer to this as the \textbf{sparse-zero bipartite setting}, which defines the key challenge central to drug repurposing for rare diseases.

Under this condition, the reliance of KGs and GNNs on message passing between neighboring nodes becomes a critical liability: associations from the target disease to a candidate drug are too weak to carry signals, limiting the model’s generalization capacity. 
One common response is to densify knowledge graphs with similarity or co-mention edges to sustain message passing~\citep{liu_drugcombdb_2020, li_deepkg_2022, chandak_building_2023}, but spurious links inevitably dilute genuine signals and obscure truly informative evidence.
As a result, it \emph{masks} the underlying scarcity and \emph{recasts} rare-disease repurposing as a conventional GNN cold-start problem, and models trained on these inflated graphs learn unreliable representations and produce many implausible candidate associations~\citep{jiang_deep_2022, yu_robust_2024, ma_learning_2024}, leaving the core challenge unresolved.

\begin{figure*}[t]
  \centering
  \includegraphics[width=1.0\textwidth]{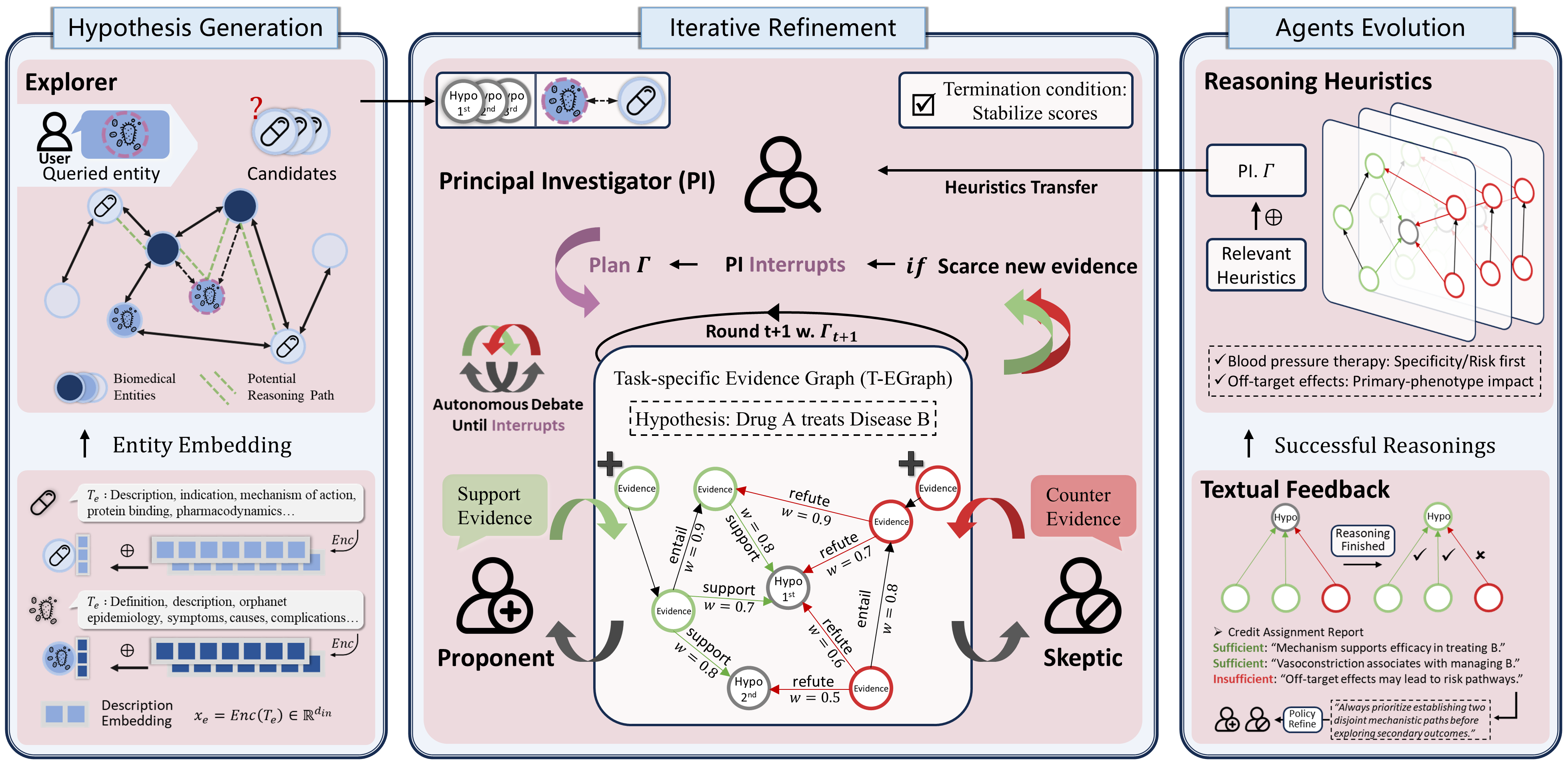}
  \caption[\model{} overview]{\model{} overview. 
  Explorer seeds a task-specific evidence graph (T-EGraph) with initial candidates based on the user query.\protect\footnotemark\ 
  The Proponent and Skeptic freely add and challenge evidence on a shared whiteboard, while the PI monitors, interrupts, and revises.
  Iteration yields ranked candidates with auditable trails.
  Finally, PI analyzes reasoning paths to generate feedback and distill heuristics, continuously improving the system's policies.}
  \label{fig:architecture}
\end{figure*}

In this paper, we advocate a paradigm shift from passive pattern recognition on static graphs to active, evidence-seeking reasoning that simulates the scientific discovery process.
We introduce \model, a self-evolving research framework designed for this challenge.
\model~casts drug repurposing for rare diseases as a dynamic investigation conducted by multiple AI agents with complementary roles: a \textbf{Principal Investigator (PI)} provides strategic oversight, an \textbf{Explorer} formulates initial hypotheses, and \textbf{Proponent} and \textbf{Skeptic} agents engage in an adversarial debate to gather and evaluate evidence.
As shown in Figure~\ref{fig:architecture}, the overall process builds on three mechanisms:

\footnotetext{``Disease$\to$Drug'' in Explorer is just an example; either disease or drug can be the query condition.}

\begin{itemize}
    \item \textbf{Evidence Reasoning as Dynamic Graph Learning.}
    To address the lack of existing relational paths, agents maintain a \textbf{Dynamic Shared Whiteboard} instantiated as a task-specific evidence graph (\textbf{T-EGraph}).
    Nodes represent the hypotheses proposed by the \textbf{Explorer}, as well as the support and counter evidence collected by the \textbf{Proponent} and \textbf{Skeptic}.
    Edges encode support, refutation, and entailment. Each action (hypothesize, gather, argue) updates the graph.
    We formulate reasoning by selecting the next action, conditioned on the current graph state, to assemble a complete evidence chain efficiently.
    
    \item \textbf{Reasoning Policy Self-Evolution via Textual Feedback.}
     In the absence of ground truth reasoning paths to learn from, \model~introduces a self-evolutionary loop driven by natural language feedback. After each T-EGraph construction, the \textbf{PI} issues a textual \textbf{Credit Assignment Report} that marks pivotal versus unproductive evidences/steps, automatically revises agent prompts, and steers future actions toward high-yield behaviors.
    
    \item \textbf{Distilling Reasoning Trajectories into Transferable Heuristics.}
    To avoid relearning common reasoning patterns for each new zero-shot case, \model~accumulates scientific intuition. \model{} meta-analyzes trajectories to extract recurring successful reasoning patterns, distills them into transferable heuristics, and stores them in a shared library of policy priors. For novel but related tasks, the \textbf{PI} initializes prompts with relevant heuristics, which narrows the search space and accelerates convergence.
\end{itemize}

\model~actively constructs plausible drug-disease hypotheses supported by a verifiable, auditable evidence trail and full reasoning path. The final output is not just a prediction but a comprehensive research report detailing the recommended pairs, the assembled body of evidence, and the complete reasoning steps leading to each conclusion, thus improving reliability and interpretability.

Our contributions can be summarized as follows:

\begin{itemize}
    \item We identify \textit{\textbf{sparse-zero bipartite setting}} as a critical yet underexplored challenge in drug repurposing for rare diseases, characterized by the absence of prior knowledge on both the drug and the target disease.
    \item We propose \model, a multi-agent research framework where agents collaboratively debate to derive and score hypotheses, grounding their reasoning in a dynamical evidence graph.
    \item We introduce a self-evolutionary learning paradigm that improves reasoning policies without direct supervision. We use an automated feedback loop of textual critiques to refine agent behaviors and distill transferable heuristics from successful reasoning trajectories.
    \item Experimental results demonstrate \model~outperforms domain-specific GNN baselines and large language models on drug repurposing tasks for rare diseases, with verifiable evidence for its predictions. Specifically, RareAgent achieves state-of-the-art results on the rare disease drug repurposing task (0.438 AUPRC / 0.662 AUROC) and on the BioHopR benchmark (17.68\% 2-hop precision).
\end{itemize}

%% file: body/3_relatedwork.tex
\section{Related Work}

\noindent\textbf{Graph-based Methods for Drug Repurposing.}
Graph neural networks (GNNs) over biomedical knowledge graphs (KGs) have become central to computational drug repurposing~\citep{perdomo-quinteiro_knowledge_2024,wei_use_2025}. Early systems such as Decagon~\citep{zitnik_modeling_2018} and DTD-GNN~\citep{li_drug_2024} established modeling paradigms for biomedical graphs, while more recent approaches like TxGNN~\citep{huang_foundation_2024} improve generalization to rare diseases via metric learning and neighborhood aggregation.
Despite these advances, performance remains coupled to graph structure and edge quality, and apparent density in KGs can be misleading. For example, in PrimeKG, 94.7\% of edges related to diseases and drugs capture under-specified drug-drug associations~\citep{chandak_building_2023}, diluting therapeutically meaningful signals and revealing limitations of GNNs when dense, high-quality relational paths are scarce~\citep{jiang_deep_2022, huang_hub-hub_2023, ma_learning_2024, jamadandi_spectral_2024}.

\noindent\textbf{LLM-powered Agents for Scientific Discovery.}
LLM-based agents integrate planning, tool use, and automated experimentation for scientific workflows. Coscientist demonstrates autonomous robotic experiments~\citep{boiko_autonomous_2023}, while ChemCrow tailors domain tools to chemistry and drug discovery. Multi-agent pipelines such as the AI Scientist series report fully AI-generated manuscripts~\citep{lu_ai_2024, yamada_ai_2025}, and systems like PaperQA enable retrieval-augmented question answering over scientific literature~\citep{m_bran_augmenting_2024}. However, these agents typically assume a well-scoped problem or some initial evidence to ground reasoning. Extending them to the \textbf{sparse-zero bipartite setting} where such grounding is absent remains a key open challenge.

\noindent\textbf{Self-improving and Evolutionary AI Systems.}
Work on self-improvement provides mechanisms for refining reasoning policies over time. Reflexion leverages verbalized reflection for reinforcement~\citep{shinn_reflexion_2023}, Self-Refine operationalizes a generate-critique-refine loop~\citep{madaan_self-refine_2023}, and evolutionary strategies such as PromptBreeder optimize prompts via self-referential mutation~\citep{fernando_promptbreeder_2023}. Complementary views frame reasoning as structured search, externalized in Tree-of-Thoughts~\citep{yao_tree_2023} and Graph-of-Thoughts~\citep{besta_graph_2024}. TextGrad formalizes language feedback as \emph{textual gradients} for systematic refinement~\citep{yuksekgonul_optimizing_2025}. A persistent frontier is to develop systems that autonomously improve from sparse, noisy, and unstructured feedback, especially in domains lacking explicit reward signals, where domain-specific intuition must be accrued gradually.

%% file: body/4_method.tex
\section{Methodology}

\begin{algorithm}[t]
\caption{\model: Iterative Refinement}
\label{alg:reasoning_loop}
\begin{algorithmic}[1]
\State \textbf{Input:} Initial Hypotheses $\{v_{h_i}\}$, Initial Graph $G_E^{(0)}$, Max rounds $T_{\max}$, Stop threshold $\delta_{\text{stop}}$
\State $\Gamma_0 \leftarrow \text{InitialPlan()}$
\For{$t = 0, 1, \dots, T_{\max}-1$}
    \State $G_E^{(t+1)} = \mathcal{M}(G_E^{(t)}, \Delta G_{E, \text{pro}}^{(t)}, \Delta G_{E, \text{ske}}^{(t)})$ 
    \Comment{Inner runs until PI interrupts (graph change $\Delta G_E < \epsilon$)}
    
    \ForAll{$v_{h_i}$}
        \State $s_{t+1}(v_{h_i}) \leftarrow \text{PI.score}(G_E^{(t+1)}, v_{h_i})$
    \EndFor
    
    \If{$\max_i |s_{t+1}(v_{h_i}) - s_t(v_{h_i})| \le \delta_{\text{stop}}$ and $t>0$}
        \State \textbf{break}
    \EndIf
    
    \State $\Gamma_{t+1} \leftarrow \text{PI.revise}(G_E^{(t+1)}, \{s_{t+1}(v_{h_i})\})$
\EndFor
\State \textbf{Output:} Ranked list $\langle (c_i, G_E^{(T)}) \rangle$ sorted by final scores $s_T(v_{h_i})$.
\end{algorithmic}
\end{algorithm}

\subsection{Problem Formulation}

A biomedical knowledge graph can be formalized as $G_K = (\mathcal{E}, \mathcal{R}, \mathcal{T})$, where $\mathcal{E}$ is the set of entities (e.g., drugs, diseases), $\mathcal{R}$ is the set of relation types, and $\mathcal{T}$ is the set of known triplets $(h, r, t)$.
A drug repurposing query is a tuple $Q = (q, r_{\text{target}})$, where $q$ is the query entity, either a drug for which we seek potential therapeutic diseases or a disease for which we seek candidate drugs, and $r_{\text{target}}$ is the target relation, such as \textit{indication} or \textit{contraindication}.
The goal is to identify a set of candidate entities $\mathcal{C} \subset \mathcal{E}$ such that the triplet $(q, r_{\text{target}}, c)$ for each $c \in \mathcal{C}$ is a valid but unobserved link.

The challenge arises when attempting to infer a relationship $(q, r_{\text{type}}, c)$ under the \textbf{sparse-zero bipartite setting}, where both endpoints lack meaningful prior evidence in the graph. 
For any entity $e \in \mathcal{E}$, its neighborhood is $\mathcal{N}(e) = \{ v \in \mathcal{E} \mid (e,r,v)\in\mathcal{T} \ \lor \ (v,r,e)\in\mathcal{T} \},$
and the sparse-zero setting for a candidate link $(q, r_{\text{type}}, c)$ is satisfied when both neighborhoods are sparse or empty:
\begin{equation}
    |\mathcal{N}(q)| \leq \epsilon_1 \quad \land \quad |\mathcal{N}(c)| \leq \epsilon_2,
\end{equation}
where $\epsilon_1,\epsilon_2$ are typically small non-negative integers (e.g., 0 or 1). 
This setting is fundamentally more challenging than conventional cold-start problems, which typically involve only one entity with sparse connections, thereby limiting the generalization ability of GNN-based models that rely on message-passing mechanisms.

Our objective under this setting is not merely to predict a score, but to generate a ranked list of candidate pairs, each accompanied by a verifiable body of evidence. The final output for a query $Q$ is a ranked list $\mathcal{C}^*$:

\begin{equation}
    \mathcal{C}^* = \langle (c_1, G_{E,1}), \dots, (c_k, G_{E,k}) \rangle
\end{equation}

where $c_i$ is a candidate entity and $G_{E,i}$ is a task-specific evidence graph (T-EGraph) that provides an auditable reasoning path justifying the proposed link $(q, r_{\text{target}}, c_i)$.

The overall objective is to learn a reasoning policy $\mathcal{\pi}$ that maps an input query to a ranked list of candidates and their evidence graphs: $\mathcal{\pi}: (q, r_{\text{type}}) \mapsto \mathcal{C}^*$. The policy $\mathcal{\pi}$ is optimized to maximize both the relevance and accuracy of the candidates $c_i$, and the logical coherence and completeness of the supporting evidence graphs $G_{E,i}$.

\subsection{RareAgent Framework}

\model~operationalizes the scientific discovery process using a team of four AI agents with specialized roles, who collaborate on a shared whiteboard by constructing T-EGraph.

\begin{itemize}

    \item \textbf{Principal Investigator (PI)}: The strategic orchestrator. The PI does not directly manipulate evidence but sets research goals, evaluates the state of the T-EGraph, identifies knowledge gaps, issues directives to other agents, and is responsible for self-evolutionary learning loops.

    \item \textbf{Explorer}: The hypothesis generator. The Explorer initiates the investigation by querying the external KG using GNN to propose an initial set of plausible drug-disease hypotheses.

    \item \textbf{Proponent}: The constructive advocate. The Proponent's goal is to build a coherent, mechanistic case for a given hypothesis. It searches for and adds supporting evidence to the T-EGraph, aiming to form complete causal chains (e.g., Drug $\rightarrow$ Target $\rightarrow$ Pathway $\rightarrow$ Phenotype).

    \item \textbf{Skeptic}: The adversarial challenger. The Skeptic's role is to challenge the current hypothesis by finding counter-evidence, identifying risks, and exposing logical fallacies. It adds refuting evidence, such as off-target effects or safety concerns, to the T-EGraph.

\end{itemize}

The overall workflow proceeds in three phases: (1) \textbf{Hypothesis Generation}, where the Explorer seeds the investigation; (2) \textbf{Iterative Refinement}, an autonomous adversarial debate where the Proponent and Skeptic build upon the T-EGraph under the PI's guidance; and (3) \textbf{Self-Evolution}, where the PI analyzes completed investigations to refine agent policies and distill reusable knowledge. All four agents are realized as LLM modules with role-specific policies, and the backbone is swappable and orthogonal to our contributions.

\subsection{Hypothesis Generation}

The \textbf{Explorer} surveys potential drug-disease connections and proposes a tractable set of initial hypotheses in the \emph{sparse-zero} setting, where usable paths in $G_K$ are absent.
It derives semantic representations from biomedical ontologies and descriptions, and applies a relation-aware scoring module to rank candidates for a query $Q=(q, r_{\text{target}})$.
We select the top-$k$ candidates to form the initial hypotheses and construct the initial T-EGraph $G_E^{(0)}$ for subsequent refinement.
Implementation and modeling details are provided in Appendix~\ref{app:explorer}.

\subsection{Iterative Refinement}
Once initial hypotheses are proposed, the framework enters a refinement loop. This phase forms the core of \model's reasoning process, where the \textbf{Proponent} and \textbf{Skeptic} agents engage in an adversarial debate, dynamically constructing a T-EGraph under the strategic guidance of the \textbf{PI}.

\noindent\textbf{Construction of the Task-specific Evidence Graph (T-EGraph).}
For a given query $Q=(q,r_{\text{type}})$, a T-EGraph is a directed typed graph $G_E = (V_E, E_E, \tau, \rho),$ where $\tau:V_E\!\to\!\{\text{query},\text{hypothesis},\text{evidence}\}$ assigns node roles and $\rho:E_E\!\to\!\{\text{entails},\text{supports},\text{refutes}\}$ assigns edge semantics.
Each edge carries an \textbf{evidence weight} $w \in [0, 1]$ assigned by the creating agent, representing the confidence in or strength of that relational claim.

\noindent\textbf{Adversarial Debate of Graph Refinement Loop.}
At each round $t$, the system state is the current T-EGraph, $G_E^{(t)}$. The loop, detailed in Algorithm \ref{alg:reasoning_loop}, consists of three steps:

(1) \textbf{Adversarial Debate (\texttt{Proponent/Skeptic:} \texttt{debate})}:
A round begins with the Proponent and Skeptic engaging in an autonomous debate.
They \textit{freely} add evidence to the T-EGraph, reacting to the current state of evidence and each other's arguments.
The agents execute their assigned \texttt{graph\_actions}.
Their policies, $\pi_{\text{pro}}$ and $\pi_{\text{ske}}$, map the current state and the directive to a graph update $\Delta G_E$. This produces the next graph state:
\begin{equation}
    G_E^{(t+1)} = \mathcal{M}(G_E^{(t)}, \Delta G_{E, \text{pro}}^{(t)}, \Delta G_{E, \text{ske}}^{(t)})
\end{equation}
where $\mathcal{M}$ is a graph manager command conducted by PI that merges updates and ensures consistency.
Notably, this process is competitive and interactive rather than a rigid turn-taking exchange.

(2) \textbf{Supervised Interruption and Scoring (\texttt{PI}: \texttt{monitor} \texttt{\&} \texttt{score})}:
The PI monitors the debate and interrupts the process if lack of substantive new evidence.
Then PI evaluates the T-EGraph for each hypothesis.
It computes a multi-faceted score $s_t(v_h)$ using an aggregation command $\mathcal{A}$ that combines structural graph properties.
The command is guided by a set of \textbf{scoring weights} that are dynamically instantiated by the PI from the available evidence, which prioritize different aspects of the evidence (e.g., placing a higher weight on mechanistic coherence than on safety signals in early rounds).
\begin{equation}
    s_t(v_h) = \mathcal{A} \left( \sum w_{\text{sup}}, \sum w_{\text{ref}}, C_{\text{mech}}, D_{\text{path}}, \dots \right)
\end{equation}
where $\sum w_{\text{sup}}$ and $\sum w_{\text{ref}}$ are the aggregated weights of supporting and refuting evidence, $C_{\text{mech}}$ is a metric for mechanistic connectivity, and $D_{\text{path}}$ measures the disjointness of evidence paths.

(3) \textbf{Strategic Revision (\texttt{PI:} \texttt{revise})}: Based on the scores, the PI generates a strategic plan $\Gamma_t$, a set of structured directives (\texttt{graph\_actions}) for the Proponent and Skeptic.
An action might instruct the Proponent to ``find a second, disjoint mechanistic path supporting H1'' or task the Skeptic to ``stress-test the risk of supine hypertension for H1''.

\noindent\textbf{Termination Condition.}
The iterative loop terminates when one of two conditions is met: (1) the hypothesis scores and rankings stabilize, i.e., the maximum change in any score falls below a predefined threshold $\delta_{\text{stop}}$, or (2) a maximum number of refinement rounds, $T_{\max}$, is reached. This ensures that the process concludes when a consensus is approached or resources are exhausted.

\input{table/rare}

\subsection{Agents Evolution}
\model~achieves long-horizon learning by distilling and reusing \emph{reasoning patterns} extracted from completed trajectories.

\noindent\textbf{Learning from Textual Feedback.}
After each complete investigation, the PI performs a final audit and generates a \textbf{Credit Assignment Report}, $R_{\text{final}}$. This is a structured natural language critique of the entire reasoning trajectory, identifying pivotal motifs (successful reasoning patterns), unproductive paths, and residual gaps. This report is then used to automatically refine the agents' policies, which are implicitly encoded in their instructions, via a refinement operation:
\begin{equation}
    P_{\text{agent}}^{(k+1)} \leftarrow \mathcal{U}(P_{\text{agent}}^{(k)}, R_{\text{final}}^{(k)})
\end{equation}
where $P_{\text{agent}}^{(k)}$ is the instruction for an agent on task $k$, and $\mathcal{U}$ is an update command executed by PI that synthesizes the feedback into concrete improvements for the agent's instructions (e.g., adding \textit{``Always prioritize establishing two disjoint mechanistic paths before exploring secondary outcomes.''}).

\noindent\textbf{Distilling and Transferring Reasoning Heuristics.}
To accumulate ``scientific intuition'' and avoid relearning common reasoning patterns, \model~distills successful strategies into a library of transferable heuristics.

(1) \textbf{Distillation}: The PI meta-analyzes a collection of high-quality Credit Assignment Reports to identify recurring successful patterns. It then abstracts them into concise, conditional rules, or \textbf{heuristics} (e.g., \textit{``WHEN disease is neurogenic, THEN prioritize agents acting on \(\alpha_1\)-receptors and check for supine hypertension risk.''}). A new heuristic $H_{\text{new}}$ is generated and added to a shared library, $\mathcal{L} \leftarrow \mathcal{L} \cup \{H_{\text{new}}\}$.

(2) \textbf{Transfer}: When a new query $Q_{\text{new}}$ arrives, the PI retrieves the top-$J$ most relevant heuristics from the library $\mathcal{L}$ and injects them into the initial prompts for the PI. This bootstrapping mechanism primes the agents with relevant prior knowledge, effectively narrowing the search space and accelerating convergence on new but related problems.

%% file: table/rare.tex
\begin{table*}[t]
  \centering
  \setlength{\tabcolsep}{4pt}
  \renewcommand{\arraystretch}{1.05}
  \begin{threeparttable}
    \resizebox{\textwidth}{!}{%
    \begin{tabular}{lcccc|cccc}
      \toprule
      \multirow{2}{*}[-0.2ex]{\makecell[l]{Method}} &
      \multicolumn{4}{c|}{\textbf{Indication}} &
      \multicolumn{4}{c}{\textbf{Contraindication}} \\
      \cmidrule(lr){2-5} \cmidrule(lr){6-9}
      & \makecell[c]{AUPRC$\uparrow$} & \makecell[c]{AUROC$\uparrow$} &
        \makecell[c]{P@10$\uparrow$}  & \makecell[c]{R@10$\uparrow$}  &
        \makecell[c]{AUPRC$\uparrow$} & \makecell[c]{AUROC$\uparrow$} &
        \makecell[c]{P@50$\uparrow$}  & \makecell[c]{R@50$\uparrow$}  \\
      \midrule
      TxGNN  & 0.373 & 0.552 & 0.02 & 0.12 & 0.241 & 0.448 & 0.00 & 0.00 \\
      \midrule
      gpt-4o-mini        & 0.361 & 0.544 & 0.01 & 0.06 & 0.244 & 0.458 & 0.00 & 0.00 \\
      gpt-4o   & 0.421 & 0.594 & 0.03 & 0.19 & 0.283 & 0.495 & 0.0012 & 0.0588 \\
      o3-mini      & 0.376 & 0.554 & 0.02 & 0.14 & 0.269 & 0.490 & 0.00 & 0.00 \\

      \midrule
      gpt-4o-mini (w.Con)        & 0.361 & 0.524 & 0.01 & 0.11 & 0.245 & 0.436 & 0.00 & 0.00 \\
      gpt-4o (w.Con)  & 0.421 & 0.599 & 0.03 & 0.20 & 0.245 & 0.443 & 0.0012 & 0.0716 \\
      o3-mini (w.Con)     & 0.432 & 0.607 & 0.03 & 0.26 & 0.260 & 0.487 & 0.00 & 0.00 \\
      
      \midrule
      ToT          & 0.371 & 0.528 & 0.01 & 0.09 & 0.251 & 0.442 & 0.00 & 0.00  \\
      GoT          & 0.206 & 0.479 & 0.01 & 0.09 & 0.260 & 0.444 & 0.0012 & 0.0196 \\
      TextGrad     & 0.138 & 0.490 & 0.00 & 0.00 & 0.241 & 0.433 & 0.00 & 0.00 \\
      \midrule
      \model   & \textbf{0.438} & \textbf{0.662} & \textbf{0.04} & \textbf{0.27} & \textbf{0.289} & \textbf{0.497} & \textbf{0.0024} & \textbf{0.0735} \\
      \bottomrule
    \end{tabular}}
    \caption{\textbf{Rare-disease drug repurposing on the 100,000 Genomes Project.}
  We compare model predictions with real-world experimental outcomes on rare-disease cohorts.
  Metrics: \emph{AUPRC}/\emph{AUROC}; \emph{P@k}, precision among the top-k predicted drugs per disease; \emph{R@k}, recall among the top-k.}
  \label{tab:rare-100kgp-repurpose}
  \end{threeparttable}
\end{table*}

%% file: body/5_experiments.tex
\section{Experiments}

In this section, we evaluate \model~through four research questions (RQs).
First, we examine drug repurposing for rare diseases under the challenging sparse-zero bipartite setting (RQ1).
Second, we assess its performance on general biomedical reasoning tasks (RQ2).
Then, we perform ablation studies to quantify the contributions of key components (RQ3).
Finally, we present a case study illustrating \model’s ability to produce auditable and interpretable evidence trails (RQ4).

\subsection{Datasets and implementation details}
We evaluate RareAgent on three datasets:
(1) 100,000 Genomes Project.
We use the disease list from the Genomes Project to enhance clinical relevance.
Diseases are from the project’s \textit{Current Rare Disease List v1.9.0}\footnote{\url{https://www.genomicsengland.co.uk/initiatives/100000-genomes-project/rare-disease}}, and ground-truth drug-disease pairs are matched from clinical trials\footnote{\url{https://clinicaltrials.gov/}}, 230 diseases and 74 drugs survive quality filters, yielding 1188 candidate pairs.
(2) PrimeKG~\citep{chandak_building_2023}.
A large, multimodal biomedical KG integrating 20 curated sources (129,375 nodes across 10 types, 4M edges) with strong rare disease coverage and natural language node descriptions. 
(3) BioHopR~\citep{kim_biohopr_2025}.
A PrimeKG-derived benchmark for multi-hop biomedical reasoning, we use it to assess generalization beyond repurposing and to compare against state-of-the-art baselines.

\subsection{RQ1: Rare disease drug repurposing}

We evaluate rare-disease cohorts from the Genomics England 100,000 Genomes Project, using registered interventional trials as the ground truth. 
We analyze RareAgent and compare against (i) GNN baseline, (ii) general-purpose LLMs, and (iii) reasoning-augmented frameworks.
Baselines marked ``w.Con'' enable self-consistency~\citep{wang_self-consistency_2022}, and all reasoning frameworks use the same base model (gpt-4o-mini) with \model~and are budget-matched in LLM call counts.
Comprehensive results are in Table~\ref{tab:rare-100kgp-repurpose}.

\noindent
(1) \textbf{Indication.} RareAgent outperforms reasoning frameworks under identical computational constraints, yielding an 18.1\% increase in AUPRC compared to ToT, highlighting the value of evolving, dynamic evidence graphs over scaffolded deliberation alone. 
(2) \textbf{Contraindication.} All methods exhibit near-zero top-K precision/recall, but RareAgent still outperforms general and reasoning baselines in ranking, underscoring the robustness of adversarial evidence aggregation when positive signals are rare.
(3) Notably, despite using gpt-4o-mini as the base, RareAgent outperforms o3-mini (w.Con): +1.4\% AUPRC, +9.1\% AUROC on Indication. This indicates that structured evidence aggregation and self-evolution can outweigh raw model capacity.

\begin{figure}[t]
  \centering
  \includegraphics[width=0.75\textwidth]{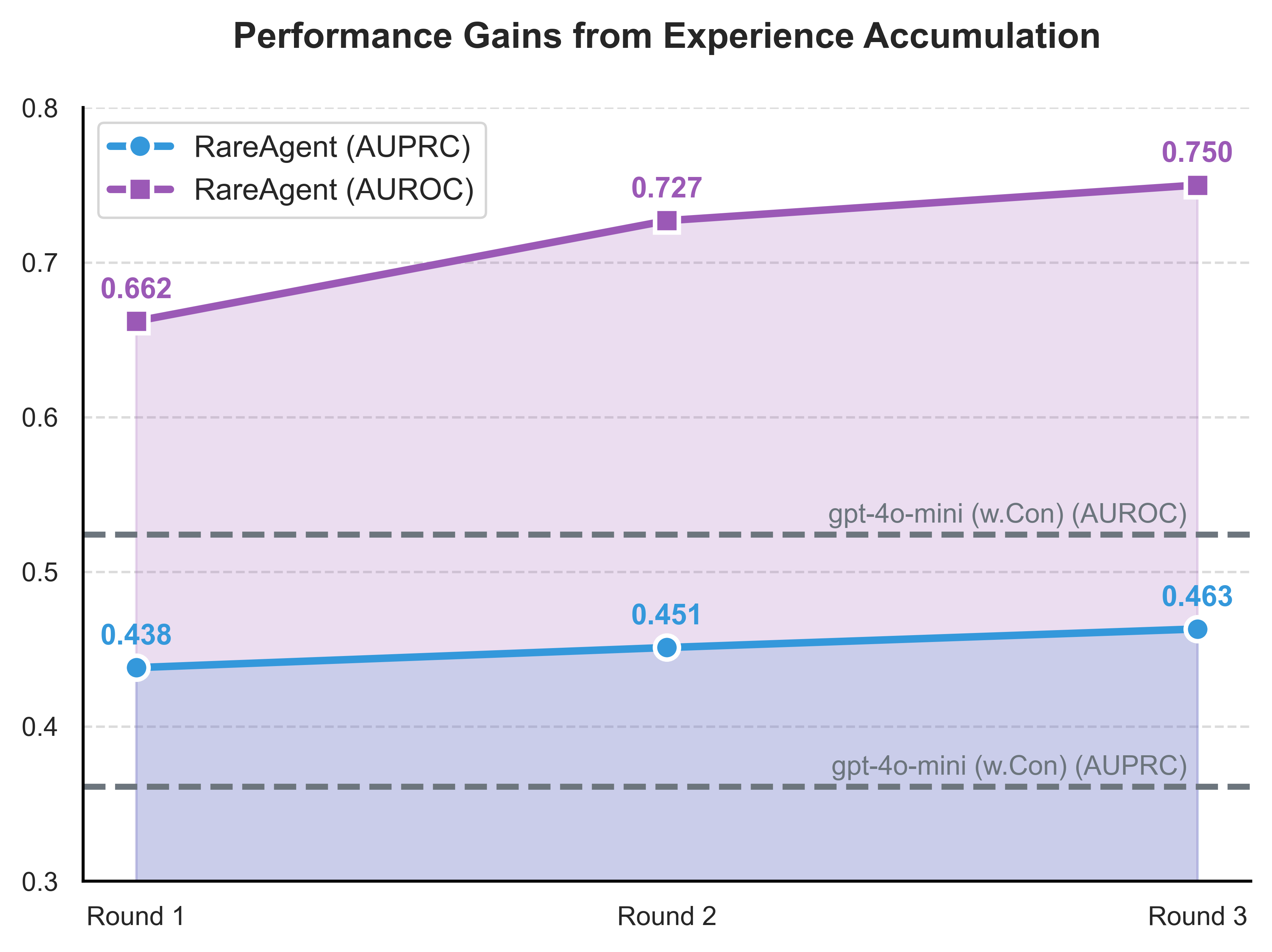}
  \caption{The cumulative effect of multiple self-evolution rounds on agent performance.}
  \label{fig:evo}
\end{figure}

To further investigate the cumulative impact of our self-evolution mechanism, we visualize RareAgent's performance across three distinct self-evolution rounds in Figure~\ref{fig:evo}. The results reveal a distinct and monotonic improvement, with AUPRC rising from 0.438 to 0.463 and AUROC increasing from 0.662 to 0.750. This steady upward trend validates that our self-evolutionary loop enables the system to effectively accumulate experience from prior investigations, thereby enhancing its long-horizon predictive accuracy.

\subsection{RQ2: General biomedical reasoning}

To assess whether the benefits observed on rare-disease repurposing transfer to broader biomedical reasoning, we evaluate on BioHopR, a PrimeKG-derived multi-hop benchmark.
We follow the original evaluation protocol and compare against strong general and medical LLM baselines. 

\noindent
(1) \textbf{Main results are shown in Table~\ref{tab:biohopr-main}}. RareAgent attains 33.87\% on 1-hop precision and 17.68\% on 2-hop precision. On 2-hop, RareAgent exceeds all baselines and indicates stronger multi-entity chain integration. On 1-hop, RareAgent using 4o-mini surpasses gpt-4o. These trends are consistent with RareAgent’s design, which constructs and refines evidence graphs through adversarial debate and self-evolution, particularly beneficial when reasoning must traverse longer causal chains. 
(2) \textbf{Relation and answer-set analysis are in Figure~\ref{fig:bio}}. Under BioHopR’s answer-split evaluation, RareAgent achieves 56.49\% / 37.22\%(Single/Multi) on 1-hop, and 21.07\% / 13.66\% on 2-hop. We attribute this pattern to T-EGraph’s emphasis on assembling mechanistically coherent paths and the PI’s directives that encourage disjoint evidence routes, which together improve coverage in stricter Multi settings.

\input{table/biohopr}

\begin{figure*}[t]
  \centering
  \includegraphics[width=1.0\textwidth]{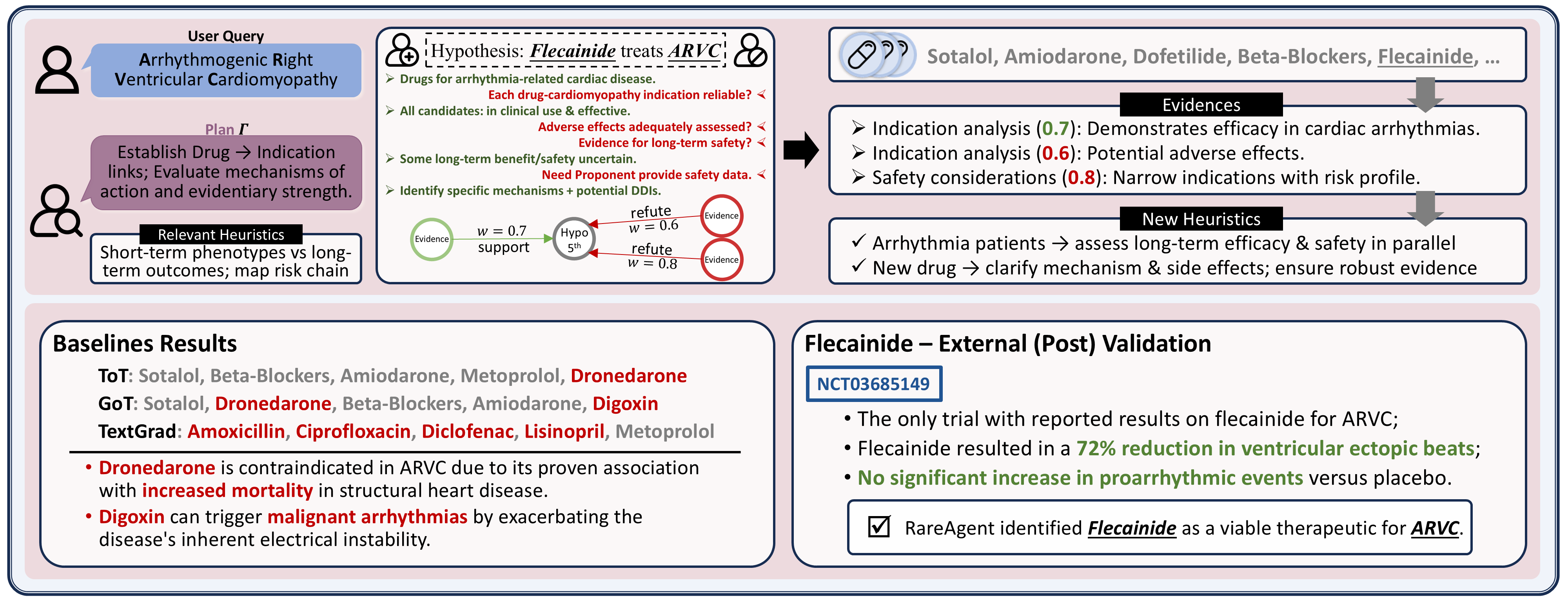}
  \caption{Case study. \model~identified \textit{\textbf{Flecainide}} as a candidate therapy for \textit{\textbf{ARVC}}, whereas the baseline found no novel therapy and proposed drugs with severe safety risks. The sole flecainide-ARVC trial supports this conclusion, and the \model~did not access that trial record during its operation.}
  \label{fig:case}
\end{figure*}

\subsection{RQ3: Ablations of key components}

We perform ablations to quantify the contribution of (i) the adversarial debate among Proponent/Skeptic, (ii) the self-evolution with textual feedback and transferable heuristics, and (iii) the termination condition.
We maintain the RQ1 data, splits, and metrics.
Results are shown in Table~\ref{tab:ablation-100kgp}.

\noindent
(1) \textbf{Self-evolution is decisive}.
Removing textual feedback causes the most significant drops: AUPRC -31.1\% and AUROC -16.1\%.
Disabling heuristic transfer also degrades performance.
These results indicate that post-hoc critique and distilled priors drive most of the ranking gains. 
(2) \textbf{Adversarial debate matters}.
Collapsing the debate to a single-agent setting yields AUPRC -7.2\%, AUROC -3.9\%, while removing only the Skeptic produces a clear hit on top-K (P@10 -14.3\%, R@10 -10.7\%). 
(3) \textbf{PI interruption/termination helps}.
When the PI does not interrupt intermediate rounds (i.e., scores only at the end), ranking quality falls.
This supports the design choice of convergence-based termination and supervised interruption to curb unproductive exploration and consolidate strong evidence paths more effectively.

\input{table/ablation}

\subsection{RQ4: Case Study: Auditable Evidence Trail for ARVC (Flecainide)}

We use Arrhythmogenic Right Ventricular Cardiomyopathy (ARVC) as a case study to demonstrate RareAgent's auditable reasoning. ARVC's combination of rarity (approx. 1/5000 to 1/2000), lack of approved drugs, and high clinical severity (a leading cause of sudden cardiac death in the young)~\citep{shah_arrhythmogenic_2025} makes it a representative stress test for drug repurposing.
As shown in Figure~\ref{fig:case}, RareAgent elevates the hypothesis that \textit{``Flecainide treats ARVC''} and subjects it to debate, where refute edges immediately contest the proposal based on safety risks.
This process distills new safety-oriented heuristics that explicitly penalize the types of dangerous candidates proposed by baseline methods, such as the contraindicated \textit{Dronedarone} and high-risk \textit{Digoxin}.
This mechanism allows RareAgent to prune unsafe reasoning paths, a crucial capability that distinguishes it from other reasoning frameworks.
The only trial with reported results for this indication (NCT03685149)\footnote{\url{https://clinicaltrials.gov/study/NCT03685149}} confirms that flecainide results in a 72\% reduction in ventricular ectopic beats, with no significant increase in proarrhythmic events, compared to placebo. Thus, RareAgent correctly identifies Flecainide as a viable therapeutic for ARVC, demonstrating coherence with real-world clinical evidence.

%% file: table/biohopr.tex
\begin{table}[t]
  \centering
  \setlength{\tabcolsep}{3pt}
  \renewcommand{\arraystretch}{1.15}
  \begin{threeparttable}
  \resizebox{0.75\columnwidth}{!}{%
    \begin{tabular}{lcccc}
      \toprule
      \makecell[l]{Method} & \makecell[c]{Prec\_HOP1\\(\%)} & \makecell[c]{Prec\_HOP2\\(\%)} & \makecell[c]{BOTH\_COR\\(\%)} & \makecell[c]{BOTH\_WR\\(\%)$\downarrow$} \\
      \midrule
      Llama-3.1-8B        & 0.12  & 0.05  & 0.00 & 99.76 \\
      HuatuoGPT-o1-70B    & 0.16  & 0.00  & 0.00 & 99.93 \\
      HuatuoGPT-o1-8B     & 0.20  & 0.04  & 0.00 & 99.54 \\
      UltraMedical-8B     & 13.75 & 5.21  & 2.28 & 82.33 \\
      Llama-3.3-70B       & 25.58 & 9.58  & 4.94 & 68.33 \\
      Llama-3.1-70B       & 26.38 & 9.47  & 4.93 & 65.64 \\
      gpt-4o-mini          & 28.11 & 14.57 & 6.54 & 64.69 \\
      gpt-4o               & 32.88 & 14.57 & 7.86 & 57.96 \\
      \addlinespace[2pt]
      \midrule
      \model~          & \textbf{33.87}     & \textbf{17.68}    &  \textbf{9.74}  & \textbf{56.52}    \\
      \bottomrule
    \end{tabular}}
    \caption{\textbf{BioHopR main results.} \emph{Prec\_HOP1/2}: precision on 1-hop/2-hop tasks; \emph{BOTH\_COR/WR}: fraction where both hops are correct/wrong.}
  \label{tab:biohopr-main}
  \end{threeparttable}
\end{table}

\begin{figure}[t]
  \centering
  \includegraphics[width=0.75\textwidth]{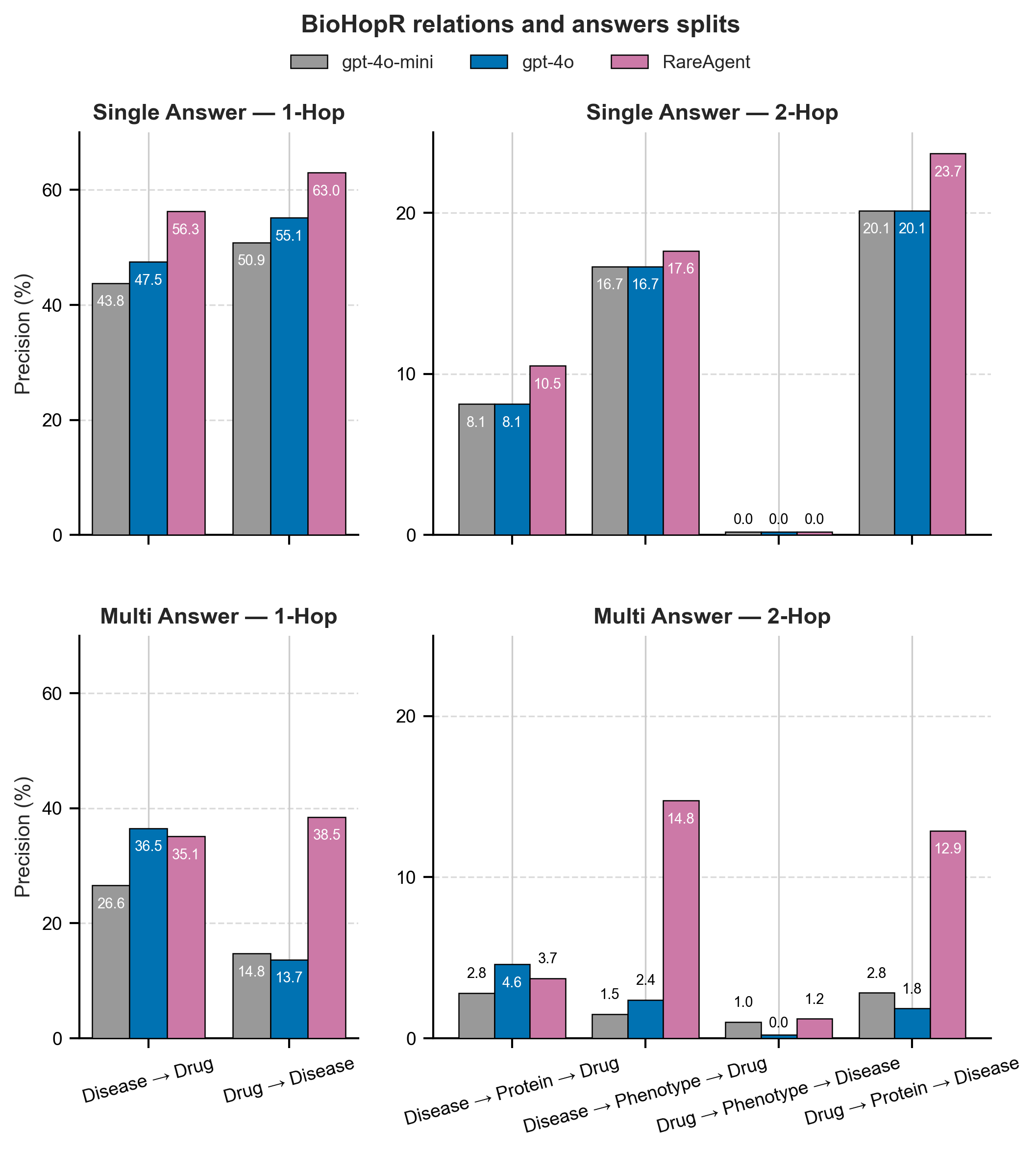}
  \caption{\textbf{BioHopR relations and answers splits.} \emph{Single} asks for one correct answer per hop; \emph{Multi} asks for all correct answers.}
  \label{fig:bio}
\end{figure}


%% file: table/ablation.tex
\begin{table}[t]
  \centering
  \setlength{\tabcolsep}{5pt}
  \renewcommand{\arraystretch}{1.15}
  \begin{threeparttable}
    \resizebox{0.75\columnwidth}{!}{%
    \begin{tabular}{lcccc}
      \toprule
      \makecell[l]{Configuration} &
      \makecell[c]{AUPRC} &
      \makecell[c]{AUROC} &
      \makecell[c]{P@10} &
      \makecell[c]{R@10} \\
      \midrule
      \model       & 0.4381 & 0.6623 & 0.0412 & 0.2745 \\
      \addlinespace[2pt]
      \midrule
      -Debate (single-agent)             & 0.4067 & 0.6364 & 0.0412 & 0.2353 \\
      -Skeptic (no critique)             & 0.4314 & 0.6570 & 0.0353 & 0.2451 \\
      -PI interrupts (final-only)        & 0.4145 & 0.6364 & 0.0412 & 0.2647 \\
      -Textual feedback               & 0.3018 & 0.5556 & 0.0412 & 0.2647 \\
      -Heuristic transfer             & 0.3877 & 0.6518 & 0.0353 & 0.2353 \\
      \addlinespace[2pt]
      \bottomrule
    \end{tabular}}
    \caption{\textbf{Ablation study.}
  We report the same evaluation protocol as in table~\ref{tab:rare-100kgp-repurpose}.}
  \label{tab:ablation-100kgp}
  \end{threeparttable}
\end{table}

%% file: body/6_conclusion.tex
\section{Conclusions}
In this work, we introduced RareAgent, a self-evolving multi-agent framework that reframes drug repurposing for rare diseases from passive pattern recognition to active, evidence-seeking reasoning. Our approach is tailored for the sparse-zero bipartite setting, where the absence of prior drug-disease associations critically limits traditional methods.
RareAgent’s novelty lies in three mechanisms: an adversarial debate to autonomously construct a task-specific evidence graph; a self-evolutionary loop that refines agent policies via automated textual feedback; and the distillation of successful reasoning trajectories into transferable heuristics. Together, these features enable the system to ground hypotheses in auditable evidence trails and accumulate domain-specific intuition to accelerate future investigations.
Comprehensive experiments demonstrate that RareAgent significantly outperforms both domain-specific GNN baselines and large language models in rare disease drug repurposing, with strong generalization in broader biomedical reasoning tasks. Ultimately, by producing verifiable and interpretable results, RareAgent offers a robust solution for complex scientific discovery in high-stakes, data-limited fields.

\newpage

\section*{Limitations}
While RareAgent demonstrates a significant advancement in drug repurposing for rare diseases, we acknowledge several limitations that warrant future investigation. The performance of RareAgent is fundamentally dependent on the capabilities of the underlying LLMs that power the agents. Although our framework is designed to be model-agnostic, the quality of the generated evidence and reasoning is directly tied to the LLM's factuality and susceptibility to hallucination. Furthermore, the mechanism for distilling and transferring reasoning heuristics, while effective for accelerating convergence on related tasks, carries a risk of overfitting. Heuristics derived from one class of diseases may not generalize effectively to novel queries involving fundamentally different biological pathways. Future work should explore methods to assess the generalizability of these learned heuristics and adapt them to more distant problem domains.


%% file: body/7_appendix.tex
\appendix
\section{Appendix}
\subsection{Explorer: Encoding and Scoring Details}
\label{app:explorer}

\paragraph{Entity Encoding.}
For every entity $e \in \mathcal{E}$, we construct a textual representation $T_e$ by concatenating its ontology and descriptive text.
A pre-trained sentence encoder $\text{Enc}(\cdot)$ produces a fixed feature embedding $\mathbf{x}_e = \text{Enc}(T_e) \in \mathbb{R}^{d_{in}}$, yielding a matrix $\mathbf{X} \in \mathbb{R}^{|\mathcal{E}| \times d_{in}}$.
These features remain frozen in subsequent training.

\paragraph{Relation-aware Projection.}
To bridge static descriptions and relational contexts, we employ type-specific projections ($\text{Proj}_{\text{drug}}$, $\text{Proj}_{\text{disease}}$) that map $\mathbf{x}_e$ to a complex embedding $(\mathbf{h}_e^r,\mathbf{h}_e^i) \in \mathbb{R}^d \times \mathbb{R}^d$.
Each relation $r \in \mathcal{R}$ is represented as $(\mathbf{r}^r,\mathbf{r}^i)$.

\paragraph{Scoring Function.}
Given a triplet $(q,r,c)$, plausibility is estimated via a complex bilinear interaction:
\begin{equation}
\begin{split}
    f(q, r, c) ={}& \langle \mathbf{h}_q^r, \mathbf{r}^r, \mathbf{h}_c^r \rangle + \langle \mathbf{h}_q^i, \mathbf{r}^r, \mathbf{h}_c^i \rangle \\
                  & + \langle \mathbf{h}_q^r, \mathbf{r}^i, \mathbf{h}_c^i \rangle - \langle \mathbf{h}_q^i, \mathbf{r}^i, \mathbf{h}_c^r \rangle .
\end{split}
\end{equation}

\paragraph{Hypothesis Set Construction.}
For a query $Q=(q, r_{\text{target}})$, the Explorer scores all valid candidates $c$ with $f(q,r_{\text{target}},c)$ and selects the top-$k$ to initialize the hypothesis set, which seeds the T\!-EGraph $G_E^{(0)}$ used in the refinement phase.

\subsection{Core Agent Prompts}
\label{app:agent-prompts}

The following are \emph{functional summaries} of the core agents' prompts used in \textsc{RareAgent}. They clarify roles, I/O contracts, and hard constraints.

\subsubsection{PI (Principal Investigator)}
\paragraph{Role} Evaluate and rank hypotheses, plan/ revise graph-building \& debate tasks, optionally request seed regeneration, and produce final reports/heuristics.

\paragraph{Modes}
\begin{itemize}
  \item \texttt{init}: Specify number of rounds, scoring weights, stopping criteria, and the overall plan.
  \item \texttt{score}: Score and rank each hypothesis using the current T-EGraph snapshot.
  \item \texttt{revise}: Issue graph actions to Proponent/Skeptic and set debate foci; optionally request Explorer to regenerate seeds (request only, no names).
  \item \texttt{report\_and\_evolve}: Produce a concise report, credit assignment, prompt patches, and distilled heuristics.
\end{itemize}

\paragraph{Common input (schematic)}~{}
\begin{lstlisting}
{
  "mode": "init|score|revise|report_and_evolve",
  "query": {"entity":"string","relation":"string"},
  "hypotheses": [{"id":"H1","candidate":{"name":"string"}}],
  "tegraph_snapshot": {"nodes":[...],"edges":[...],"round_index":1},
  "history": {"round":1,"last_scores":[...]},
  "thresholds": {"stop_delta":0.03,"saturation_ratio":0.65},
  "seed_context": {"target_type":"drug|disease","seed_history":[["..."]]}
}
\end{lstlisting}

\paragraph{Typical outputs}
\begin{itemize}
\item \textbf{\texttt{init}}:
\begin{lstlisting}
{"plan":{"rounds":2,"stopping":{"delta_threshold":0.03}}}
\end{lstlisting}

\item \textbf{\texttt{score}}:
\begin{lstlisting}
{
  "scoring_summary":[{"hypothesis_id":"H1","score":0.68}],
  "ranking":["H1","H2"],
  "delta_since_last_round":0.04,
  "stop_decision":{"should_stop":false}
}
\end{lstlisting}

\item \textbf{\texttt{revise}}:
\begin{lstlisting}
{
  "revisions":[
    {
      "hypothesis_id":"H1",
      "graph_actions":[
        {"type":"add_mechanism_link","assignee":"Proponent"}
      ],
      "debate_focus":["mechanistic closure & safety hotspots"]
    }
  ],
  "seed_request":{"should_regenerate":false,"reason":"example"}
}
\end{lstlisting}

\item \textbf{\texttt{report\_and\_evolve}}:
\begin{lstlisting}
{
  "final_recommendations":[{"hypothesis_id":"H1","score":0.75}],
  "prompt_patches":[{"role":"Proponent","patch":"example"}]
}
\end{lstlisting}
\end{itemize}

\paragraph{Scoring signals} Mechanism fit, class prior, PK/PD feasibility, indication plausibility, safety risk, structural graph bonus (disjoint paths/connectivity), conflict penalty.

\paragraph{Hard constraints}
\begin{itemize}
\item PI does \emph{not} directly modify the graph; it emits instructions for other agents.
\item Score every input hypothesis. Seed regeneration can only be \emph{requested}, not generated here.
\end{itemize}

\subsubsection{Proponent (Support Builder)}
\paragraph{Role} Build \emph{supportive} mechanistic chains and graph updates for a given hypothesis, or execute PI-assigned graph actions.

\paragraph{Modes}
\begin{itemize}
  \item \texttt{build\_chain}: Complete supportive paths based on common sense and heuristics.
  \item \texttt{execute\_actions}: Execute \texttt{graph\_actions} (assignee=Proponent).
\end{itemize}

\paragraph{Input (schematic)}~{}
\begin{lstlisting}
{
  "mode":"build_chain|execute_actions",
  "query":{"entity":"string","relation":"string"},
  "hypothesis":{"id":"H1","candidate":{"name":"string"}},
  "tegraph_snapshot":{"nodes":[...],"edges":[...]},
  "graph_actions":[{"type":"add_mechanism_link","assignee":"Proponent"}],
  "constraints":{"require_disjoint_paths":2}
}
\end{lstlisting}

\paragraph{Output (schematic)}~{}
\begin{lstlisting}
{
  "graph_updates":{
    "add_nodes":[
      {"id":"n1","type":"Target","label":"..."},
      {"id":"n2","type":"Pathway","label":"..."}
    ],
    "add_edges":[
      {"source":"drugX","target":"n1","relation":"acts_on","weight":0.8,"rationale":"brief"}
    ],
    "merge":[{"keep":"n2","remove":"n2_dup"}]
  },
  "subconclusions":[{"id":"C1","text":"closure formed","confidence":"medium"}],
  "uncertainties":["e.g., BBB unknown"]
}
\end{lstlisting}

\paragraph{Key constraints}
\begin{itemize}
\item Aim for $\geq$1--2 \emph{disjoint} paths: Drug $\rightarrow$ Target/Pathway $\rightarrow$ Phenotype/Disease.
\item Each new edge has a weight $[0,1]$ and a one-line rationale; merge synonyms/duplicates.
\item Do not build refutation/safety chains (that is Skeptic's role).
\end{itemize}

\subsubsection{Skeptic (Counter-Builder)}
\paragraph{Role} Construct \emph{refutation/risk} chains and conflict hotspots; or execute PI-assigned risk actions.

\paragraph{Modes}
\begin{itemize}
  \item \texttt{build\_counterchain}: Extend refutation/risk mechanisms.
  \item \texttt{execute\_actions}: Execute \texttt{graph\_actions} (assignee=Skeptic).
\end{itemize}

\paragraph{Input (schematic)}~{}
\begin{lstlisting}
{
  "mode":"build_counterchain|execute_actions",
  "query":{"entity":"string","relation":"string"},
  "hypothesis":{"id":"H1","candidate":{"name":"string"}},
  "tegraph_snapshot":{"nodes":[...],"edges":[...]},
  "graph_actions":[{"type":"stress_test_safety","assignee":"Skeptic"}]
}
\end{lstlisting}

\paragraph{Output (schematic)}~{}
\begin{lstlisting}
{
  "graph_updates":{
    "add_nodes":[{"id":"k1","type":"Pathway","label":"RiskPathway"}],
    "add_edges":[
      {"source":"drugX","target":"k1","relation":"involved_in","weight":0.7,"rationale":"brief"},
      {"source":"k1","target":"H1","relation":"refutes","weight":0.85,"rationale":"safety conflict"}
    ],
    "conflict_hotspots":[
      {"topic":"supine_hypertension","pro_nodes":["..."],"con_nodes":["..."]}
    ]
  },
  "counterclaims":[{"id":"K1","text":"directional conflict/insufficient PK","confidence":"medium"}],
  "falsification_tests":["minimal falsifiable checks..."]
}
\end{lstlisting}

\paragraph{Key constraints}
\begin{itemize}
\item Prioritize safety/contraindication, mechanistic contradiction, PK/PD barriers, and phenotype-vs-outcome gaps.
\item Use \texttt{refutes/contradicts} edges to the hypothesis/claims; annotate conflict hotspots.
\item Control node/edge growth.
\end{itemize}

\subsubsection{Shared Constraints (All Agents)}
\begin{itemize}
\item Return a single UTF-8 JSON object (no Markdown or prose).
\item No fabricated sources; no clinical advice; mark uncertainty when evidence is thin.
\item Only Proponent/Skeptic update the T-EGraph; PI never edits the graph directly.
\item Field names follow the schemas above; missing optional fields must be tolerated.
\end{itemize}